\documentclass[journal]{IEEEtai}
\usepackage{setspace}
\usepackage{microtype}
\usepackage{graphicx}
\usepackage{subcaption}
\usepackage{booktabs} 
\usepackage{multirow}
\usepackage{comment}
\usepackage{hyperref}

\usepackage{amsmath}
\usepackage{amssymb}
\usepackage{mathtools}
\usepackage{amsthm}
\usepackage{tikz}
\usepackage{pgfplots}
\usepackage{pgfplotstable}
\pgfplotsset{compat=newest}

\usepackage[capitalize,noabbrev]{cleveref}

\usepackage{algorithm}
\usepackage{algorithmicx}
\usepackage{algpseudocode}

\newcommand{\hl}[1]{\textcolor{black}{#1}}
\newcommand{\IEEEAcceptedNotice}{%
  \begingroup
  \renewcommand{\thefootnote}{}%
  \footnotetext{%
    \textcopyright~2026 IEEE.
    Personal use of this material is permitted.
    Permission from IEEE must be obtained for all other uses, in any current or future media, including reprinting/republishing this material for advertising or promotional purposes, creating new collective works, for resale or redistribution to servers or lists, or reuse of
    any copyrighted component of this work in other works.
    
    This is the accepted version of:
    A.~Kaimakamidis, I.~Mademlis, and I.~Pitas,
    ``Collaborative Knowledge Distillation via a Learning-by-Education Node Community'',
    \emph{IEEE Transactions on Artificial Intelligence},
    doi: \href{https://doi.org/10.1109/TAI.2026.3704336}{10.1109/TAI.2026.3704336}. This article has supplementary downloadable material available at the publisher's Web page.
  }%
  \endgroup
}
\begin{document}

\title{Collaborative Knowledge Distillation via a Learning-by-Education Node Community}

\author{Anestis Kaimakamidis, Ioannis Mademlis, \IEEEmembership{Senior Member, IEEE}, and Ioannis Pitas, \IEEEmembership{Life Fellow, IEEE}
\thanks{Submitted for review on 30 September 2024. This work was supported by European Union's Horizon 2020 Research and Innovation programme under Grant Agreement No. 951911 (AI4Media). \textit{Corresponding author: Ioannis Mademlis.}}
\thanks{Anestis Kaimakamidis is with the Aristotle University of Thessaloniki, Thessaloniki, 54124, Greece (e-mail: akaimak@csd.auth.gr).}
\thanks{Ioannis Mademlis is with the Aristotle University of Thessaloniki, Thessaloniki, 54124, Greece (e-mail: imademlis@csd.auth.gr).}
\thanks{Ioannis Pitas is with the Aristotle University of Thessaloniki, Thessaloniki, 54124, Greece (e-mail: pitas@csd.auth.gr).}
}



\maketitle
\IEEEAcceptedNotice
\begin{abstract}
    A novel Learning-by-Education Node Community framework (LENC) for Collaborative Knowledge Distillation (CKD) is presented, which facilitates continual collective learning through effective knowledge exchanges among diverse deployed Deep Neural Network (DNN) peer nodes. These DNNs dynamically and autonomously adopt either the role of a student, seeking knowledge, or that of a teacher, imparting knowledge, fostering a collaborative learning environment. The proposed framework triggers knowledge transfer via autonomous teacher discovery and stream-driven DNN distillation as needed, while enhancing their learning capabilities and promoting their collaboration. LENC addresses the challenges of handling diverse training data distributions and the limitations of individual DNN node learning abilities. \hl{It enables the exploitation of selected peer-teacher knowledge upon learning a new task and mitigates catastrophic forgetting in DNN nodes.} \hl{Additionally, it supports task-boundary-free continual adaptation in distributed settings via autonomous role assignment and modular forgetting mitigation, as DNN nodes receive no explicit task-boundary metadata during deployment.} Experimental evaluation on a proof-of-concept implementation demonstrates the LENC framework's functionalities and benefits across multiple DNN learning and inference scenarios. The conducted experiments showcase its ability to gradually improve the average test accuracy of the community of interacting DNN nodes in image classification problems, by appropriately leveraging the collective knowledge of all node peers. The LENC framework achieves strong performance in on-line unlabelled CKD.
\end{abstract}

\begin{IEEEImpStatement}
Collaborative Knowledge Distillation (CKD) and Continual Learning (CL) are critical in advancing the capabilities of deployed Deep Neural Networks (DNNs) in a variety of real-world scenarios, such as autonomous systems and intelligent networks. Traditional DNNs face limitations in individual learning abilities and struggle with the challenge of task-agnostic CL, often leading to issues like catastrophic forgetting. The Learning-by-Education Node Community (LENC) framework introduced in this paper addresses these challenges by creating a dynamic, collaborative environment where DNN nodes autonomously act as students or teachers. This fosters an environment of continual, bidirectional knowledge transfer and enhances the learning potential of each node. LENC's ability to facilitate task-agnostic CL and enable flexible, on-the-fly role adaptation mirrors human collaborative learning dynamics, pushing the boundaries of current CKD approaches. Through this innovative framework, the paper contributes a significant advancement in enhancing the autonomy and adaptability of DNNs, potentially transforming how these networks can be deployed and continuously improved in diverse, real-world applications.
\end{IEEEImpStatement}

\vspace{-1em}
\section{Introduction}
Deep Neural Networks (DNNs) have advanced by modeling the human brain at a very high level of abstraction. A key sociobiological trait is transactional knowledge exchange, based on teacher–student relations, which is a foundational dimension of human education systems. However, ``DNN knowledge" differs from human knowledge; it is a function $y = f(\mathbf{x};\mathbf{\theta})$ that can approximate unknown functions $y = \phi(\mathbf{x})$ by exploiting large training datasets $\mathcal{D} = \left\{ \left(\mathbf{x}_i, y_i \right), i=1,\dots, N\right\}$. As researchers can abstractly simulate both human cognition and social organization, a notable trend has emerged in the form of teacher-student machine learning methods. This transformative approach has shown great potential in how machines acquire knowledge across a wide spectrum of artificial intelligence (AI) problems \cite{ de2022structural}.

Most teacher–student communities focus on Reinforcement Learning \cite{nikonova2023efficient}, which is {out of the scope of} this article. Only a few existing methods address multi-node supervised/semi-supervised scenarios (e.g., \cite{liu2024wisdom}). In this latter context, \emph{Collaborative Knowledge Distillation} (CKD) taps the diverse knowledge of multiple DNNs to augment the community's collective knowledge \cite{zhang2018deep, guo2020online, yao2020knowledge}.

Although general KD and CL have previously been combined \cite{LiTNNLS_KDCLSurvey, SzatkowskiWACV2024_AdaptTeacher}, the situation is different for CKD. Existing CKD frameworks cannot acquire new tasks on-line: nodes remain limited to predefined tasks, and the sole CKD method that employs \emph{Continual Learning} (CL) for dynamic acquisition of novel tasks allows only a one-off GAN-to-VAE distillation \cite{ye2021lifelong}. The non-task-agnostic nature of these frameworks also limits their potential. The DNN students in such frameworks acquire knowledge only when being aware of the task boundaries, i.e., when the switch of training data from an old task to a new one is fully defined and known \cite{zhang2018deep, guo2020online, yao2020knowledge, liu2024wisdom}. In classification problems, a \emph{task boundary} is defined as the last data point of the previous task learned before the first data point of the new task to be learned. Finally, the performance of existing CKD frameworks drops significantly when only a data subset is available during training, which is closer to a real-world scenario. Consequently, they cannot operate in a dynamic realistic environment, where raw unlabelled data is the only input.

The above limitations of current CKD methods have \emph{motivated} our research. Aiming to address them, this article introduces the Learning-by-Education Node Community (LENC) framework, where nodes autonomously assume teacher or student roles in task-agnostic, on-line, transactional exchanges. Each node self-assesses its familiarity with current streaming inputs; if they are unknown, this node dynamically becomes a student and broadcasts the stream to its peers in order to detect potential teachers. If a teacher is found, its relevant knowledge is transmitted to the current student via neural distillation \cite{hinton2015distilling}, while CL is employed for retaining prior knowledge; all fully automatically.

The flexibly decided teacher/student role, as well as the task-agnostic nature of the framework, simulates a community of human nodes where an unknown data stimulus coming from the environment triggers the urge to attain relevant knowledge from other humans, e.g., from a specialized teacher. Among multi-node CKD protocols, LENC uniquely couples task-agnostic CL with dynamic role assignment, which is a key characteristic of human communities.


Without loss of generality, the description of the LENC framework below assumes a classification application. Experimental evaluation on a proof-of-concept implementation demonstrates the LENC functionalities across multiple scenarios. The conducted experiments showcase the LENC framework's ability to gradually improve the average test accuracy of the community of interacting DNN nodes in image classification problems, as well as its ability to learn on-line from small batches of unlabelled data. LENC performs well in the realistic task-agnostic, on-line CKD setting.
\vspace{-0.2em}
\section{Related Work}
The concept of teacher-student learning emerged from the idea of a single, untrained student DNN distilling the knowledge held by one or an ensemble of pretrained teacher DNNs \cite{bucilua2006model}, to strike a balance between efficiency and performance \cite{bar2019robustness}. LENC is a novel multi-node framework for CKD where multiple peer DNNs can dynamically act either as teachers or students during deployment. They autonomously select their current role based on: a) whether each one is pretrained or not, and b) whether each of them recognizes each current external test input as known or not. To achieve its goals, LENC integrates Knowledge Distillation (KD), Out-of-Distribution detection (OOD) and CL into a common framework. Below, the state-of-the-art in each of the involved areas is surveyed, while positioning LENC with respect to competing methods.
\vspace{-0.2em}
\subsection{Knowledge Distillation}
The method in \cite{li2014learning} exploits a complex pretrained DNN as a teacher in classification tasks, by using its output distribution as pseudo-labels for unlabelled data utilized in student DNN training. Notably, \cite{hinton2015distilling} demonstrated remarkable results by distilling knowledge from an ensemble of DNN teacher models into a single student model. Building upon \cite{hinton2015distilling} and \cite{li2014learning}, later studies explored teacher-student interactions to enhance KD, resulting in student DNNs exhibiting strong classification performance \cite{ye2021lifelong}, \cite{bar2019robustness}, \cite{romero2014fitnets} \cite{yim2017gift}, \cite{komodakis2017paying}, \cite{zhou2018rocket}. For example, Relational Knowledge Distillation (RKD) \cite{park2019relational} leverages structural information outputs from different teachers to refine knowledge transfer, while Curriculum Temperature for Knowledge Distillation (CTKD) \cite{li2023curriculum} dynamically controls the difficulty level of tasks during a student model's learning trajectory, by incorporating a dynamic, learnable parameter.

Recent advances have ventured beyond the classical teacher–student paradigm, introducing contrastive and on-line collaborative approaches that enrich training and improve generalization. Methods such as Complementary Relation Contrastive Distillation reframe KD as a structural feature transfer problem, using an anchor-based contrastive loss to align sample representations and inter-sample relations, thus demonstrating strong effectiveness across benchmarks \cite{zhu2021crcd}. In parallel, on-line collaborative approaches like Peer Collaborative Learning (PCL) eliminate reliance on a fixed pretrained teacher by training multiple student peers jointly in a multi-branch network; the ensemble and temporal mean of peers serve as dynamic teachers, enabling one-stage, end-to-end KD with strong generalization \cite{wu2021peer}. Together, these methods underscore a shift toward more interactive and flexible KD frameworks that generate rich representations across domains.
\vspace{-0.2em}
\subsection{Collaborative Knowledge Distillation}
In collaborative learning, teachers and students mutually instruct and learn from each other, with peer DNNs possessing either identical or diverse architectures. Notably, \cite{zhang2018deep} introduced Deep Mutual Learning (DML) for on-line KD, focusing on exchanging response-based knowledge between peer DNNs. Similarly, on-line Knowledge Distillation via Collaborative Learning (KDCL) was introduced in \cite{guo2020online}, leveraging an ensemble of soft-output activations to transfer knowledge between peer DNNs. DML was expanded in \cite{yao2020knowledge} with Dense Cross-layer Mutual distillation (DCM), enabling collaborative training of both teacher and student DNNs from the ground up. Additionally, \cite{hou2017dualnet} proposed a feature fusion method, known as DualNet, which combines features from two identically structured peer DNNs using a ``SUM" operation. In contrast, a mutual KD method via Feature Fusion Learning (FFL) in \cite{kim2021feature} aims to collaboratively learn a robust classifier by merging features from various peer DNNs, which may have different architectures.

More recent algorithms include On-the-fly Native Ensemble (ONE) \cite{zhu2018knowledge}, which is a learning strategy for multi-branch students. The DNN branches (peers) exchange knowledge to enhance the model's generalization abilities. Peer Collaborative Learning (PCL) \cite{wu2021peer} introduces a CKD approach for multi-branch students, which addresses the problem of missing the discriminative information among feature representations of DNN peers (branches). Weighted Mutual Learning (WML) \cite{zhang2022weighted} further improves the performance of the branches' ensemble by estimating each branch's relative importance. Switchable On-line Knowledge Distillation (SwitOKD) \cite{qian2022switchable} uses a dynamic threshold to switch the mode of the DNN teacher from frozen weights (``expert") to unfrozen weights (``learning") and vice versa.
\vspace{-1em}
\subsection{Continual Learning}
\label{ssec:CL}
To enable a single DNN to continuously acquire new skills without forgetting old ones, CL is essential \cite{silver2013lifelong}\cite{ring1997child}. The main hurdle is integrating new tasks \cite{pentina2015lifelong} without ``catastrophic forgetting" of prior knowledge \cite{french1999catastrophic}\cite{mccloskey1989catastrophic}. Regularization approaches tackle this by penalizing changes to important parameters.

For example, Elastic Weight Consolidation (EWC) \cite{kirkpatrick2017overcoming} assigns each DNN parameter an update penalty, based on its importance for previously learned tasks, using the Fisher Information Matrix $\mathbf{F}$. Thus, DNN adaptation uses the following regularizer:
\begin{equation}
    \mathcal{L}_c(\theta) = \sum_i \frac{\lambda}{2}F_{ii}(\theta_i - \theta^*_{o, i})^2,
\label{loss_cont}
\end{equation}
\noindent where hyperparameter $\lambda$ sets the importance of previous tasks compared to the new ones and $i$ indexes each DNN parameter. Any change to the $i$-th DNN parameter is penalized by a factor proportionate to the $i$-th diagonal entry of $\mathbf{F}$, evaluated once on the parameter set where training on the old task had converged. Details about EWC are in the Supplementary Material.

Learning without Forgetting (LwF) \cite{li2017learning} preserves old-task performance by distilling outputs from the DNN's own previous version while training on new data, without requiring access to old training data. The method deviates from vanilla KD; the teacher is the older version of the student itself and not a different DNN. Encoder-Based Lifelong Learning similarly uses autoencoders to regularize feature changes alongside a KD loss, with an initial warm-up phase \cite{rannen2017encoder}.

Replay-based CL methods instead store a small set of exemplars from past tasks and train jointly on new data and these exemplars. Incremental Classifier and Representation Learning (iCaRL) selects exemplars via a herding strategy to match class means in feature space \cite{rebuffi2017icarl}, while End-to-End Incremental Learning (EEIL) adds a balanced finetuning stage with equal exemplar counts per class \cite{castro2018end}. Since the resulting dataset is skewed toward new data – causing task-recency bias – methods like Learning a Unified Classifier Incrementally via Rebalancing (LUCIR) introduce cosine normalization, a less-forget constraint, and inter-class terms to rebalance learning \cite{hou2019learning}, and Pooled Outputs Distillation Network (POD-Net) builds on LUCIR by applying pooled-output KD with a local similarity classifier \cite{douillard2020podnet}.
\vspace{-2em}
\subsection{Task-Agnostic Continual Learning}
\label{ssec::Related_TACL}
Task-agnostic supervised CL does not require explicit information regarding the task boundaries, i.e., at which incoming training data point the previous set of classes (task) switches to a new one. Bayesian Gradient Descent (BGD) \cite{zeno2018task} employs an on-line version of variational Bayes, updating the mean and variance for each parameter using the posterior distribution from the previous task as a prior distribution for the new task, mitigating catastrophic forgetting. However, it relies on the ``labels trick" which does break the task-agnostic assumption, since the task identity is inferred from the class labels during training. An alternative called iTAML \cite{rajasegaran2020itaml} uses meta-learning to maintain generalized parameters for all tasks, adapting to a new task with a single update at inference. Despite supporting undefined task boundaries, iTAML does require task boundaries to be known per data point during training. In contrast, Hybrid generative-discriminative Continual Learning (HCL) \cite{kirichenko2021task} models task and class distribution with a normalizing flow model, using anomaly detection for automatic task identification and combining generative replay and functional regularization to prevent catastrophic forgetting. Continual Neural Dirichlet Process Mixture (CN-DPM) \cite{lee2019neural} is an expansion-based method that allocates new resources for learning new data, formulating the task-agnostic problem as on-line variational inference of Dirichlet process mixture models. It employs neural experts, each handling a data subset, equipped with short-term memory for managing new data points and creating new experts when needed. Finally, the Task-Agnostic Continual Learning using Multiple Experts (TAME) \cite{zhu2022tame} algorithm uses multiple \emph{task experts}, which are completely separate DNNs. TAME automatically detects shifts in data distribution by keeping track of the loss value of the expert being trained, thus initializing a new expert on-line upon a high loss value variation, which potentially marks the onset of a new task. 
\vspace{-1em}
\subsection{Out-of-Distribution Detection}
\label{ssec: OOD}
To enable a DNN node to self-assess whether a test sample $\mathbf{x}$ originates from its training distribution \hl{$p(\mathbf{x}\mid\mathcal{D})$}, OOD detectors are employed during deployment to reject inputs whose predictions may be unreliable. Modern OOD detectors fall into classification-, generative-, distance-, or reconstruction-based approaches. Classification methods reshape the semantic label space into hierarchical taxonomies and apply top-down classification or group-softmax training to detect novel inputs \cite{linderman2023fine}. Alternatively, vision–language techniques learn dense label embeddings and flag inputs as ``novel" when the model's regression output lies beyond a predefined distance threshold to those embeddings \cite{fort2021exploring}.

Distance-based approaches assume that OOD samples lie far from class centroids in feature space, using metrics such as cosine similarity \cite{zaeemzadeh2021out}, Radial Basis Function kernels \cite{van2020uncertainty}, Euclidean distance \cite{huang2020feature}, or geodesic distance \cite{gomes2022igeood}. Other variants measure the feature norm in the orthogonal complement of the principal subspace \cite{wang2022vim} or learn hyperspherical embeddings to maximize inter-class angular separation and minimize intra-class dispersion (CIDER) for clearer ID/OOD separation \cite{ming2022cider}. In the above cases, the OOD detector employs the main DNN model features (activations). Reconstruction methods leverage instead the higher reconstruction error of Autoencoders on OOD inputs, including denoising variants like MoodCat \cite{yang2022out} and hybrid schemes such as Reconstruction Error Aggregated Detector (READ), which maps pixel-level reconstruction errors into the classifier's latent space \cite{jiang2023read}. A notable hybrid is asked Image Modeling for Out-Of-Distribution Detection (MOOD), which treats Masked Image Modeling (a self-supervised denoising pretext task) as a proxy for reconstruction-based distance detection \cite{li2023rethinking}.

Deep generative methods, such as flow models \cite{jiang2021revisiting}, train a separate DNN to probabilistically model the ID data distribution and thus allow test-time classification of ID and OOD data points. Instead of raw likelihood scores, likelihood ratios can be used \cite{ren2019likelihood} to avoid the occasional assignment of high likelihood to OOD data points \cite{nalisnick2018deep, choi2018waic, kirichenko2020normalizing}. Given that likelihood demonstrates a strong bias towards input complexity \cite{serra2019input}, Likelihood Regret (LR) \cite{xiao2020likelihood} has been proposed as an efficient OOD score when using a Variational AutoEncoder (VAE) to model the ID data distribution. LR is the logarithmic ratio between the likelihood obtained by the posterior distribution optimized separately for a given input and the likelihood approximated by the VAE:
\begin{equation} \label{eq: lr_calculation}
LR_\tau(\mathbf{x})= L_\tau(\mathbf{x}; \mathbf{\theta^{*}},\widehat{p}(\mathbf{x}))-L_\tau(\mathbf{x}; \mathbf{\theta^{*}}, \mathbf{\phi^{*}}),
\end{equation}
\noindent where \(\widehat{p}(\mathbf{x})\) is the optimal posterior distribution of the Encoder parameters given the input data \(\mathbf{x}\), \(\theta^{*}\) is the optimal VAE Decoder parameters obtained from the training dataset and $\phi^{*}$ is the optimal VAE encoder parameters obtained from the training dataset. LR relies on the intuition that a well-trained VAE model, when provided with a single ID test data point will exhibit only marginal improvement in likelihood if its current configuration is replaced with the one after optimization, resulting in small LR values. Instead, when presented with an OOD test data point, the model's current configuration is expected to deviate significantly from the one after optimization, leading to large LR values, because the model has not been exposed to similar data points during its training phase.

Notably, OOD detection has not been previously integrated with CKD and CL simultaneously, with the only exception being HCL \cite{kirichenko2021task} (see Section \ref{ssec::Related_TACL}).

\vspace{-1em}
\subsection{KD in Advanced Node Communities}
A few multi-node KD frameworks for supervised or semi-supervised learning have been proposed in recent years and do resemble the LENC framework in certain respects \cite{soltoggio2024collective}. Particularly, Lifelong Learning Teacher-Student (LTS) \cite{ye2021lifelong} uses Generative Adversarial Networks (GANs) as teachers to replay previously learned knowledge to their students, to avoid catastrophic forgetting. Unlike LENC, LTS supports only a fixed teacher/student role for each DNN: once a student is trained, it cannot effectively transfer its acquired knowledge to a different student. Additionally, LTS does not operate in a task-agnostic manner, i.e., the boundaries of each task are required to be known. An alternative framework called DiverseDistill \cite{liu2024wisdom} focuses on transferring the knowledge of multiple, potentially diverse Foundation Models to smaller students, but does not incorporate CL. Therefore, only untrained students are supported. Similarly, \cite{ye2024knowledge} introduces a multi-teacher ensemble feature KD framework that attempts to find the optimal feature transformations/weights before KD. Applying such methods in recommender systems, Pre-trained Recommendation Model Knowledge Distillation (PRM-KD) \cite{sun2024distillation} maps the output of multimodal teachers using a consistent scoring strategy, which is then used to distill their knowledge to a single student. Additionally, Multiple Teachers Model for detecting and Ranking (MulTMR) \cite{vitiugin2024multilingual} uses a multi-teacher weighted feature KD loss to transfer the knowledge of multiple multilingual Large Language Model (LLM) teachers into a single student of identical architecture. Data-Agnostic Consolidation (DAC) \cite{carta2024projected} introduces a data-agnostic approach to the general CL problem, using augmented OOD samples and KD to learn from new unlabelled data. In contrast to LENC, DAC focuses on CL and is not suitable for environments with multiple potential teachers or time-varying data. This also holds for the KD-based Complementary Calibration (CoCa) CL method \cite{ji2022complementary}, although CoCa introduces an interesting approach to exploit latent-space features for CL on labeled data streams.

In short, among multi-node CKD protocols, LENC uniquely combines CKD, CL and autonomous teacher discovery in a task-agnostic manner. LENC participants can be dynamically assigned teacher/student roles and exchange knowledge entirely autonomously and in perpetuity, while being deployed, across multiple independent tasks and with unlabelled, raw test data points as their only input. LENC generalizes and extends the functionalities of existing advanced node communities, minimizing the need for manual human intervention even in the presence of multiple unknown tasks.

\vspace{-1em}
\section{LENC Framework Architecture}
\label{sec::Method}
The proposed novel LENC framework defines the protocol for LENC nodes to learn tasks from other peer LENC nodes that participate in a LENC community. Consequently, every deployed LENC node can assess on-line its knowledge of incoming external test data points. In case of ignorance, it can decide to transfer it to other nodes via teacher-student interaction to: a) identify potential teachers, and b) learn from them. The integration of CL and CKD, combined with the ability of each node to self-assess its knowledge, emulates a human community where all nodes cooperate to broaden their knowledge on multiple tasks.
\vspace{-1em}
\subsection{LENC Node Architecture}
\begin{figure}
    \centering
    \includegraphics[width=0.5\linewidth]{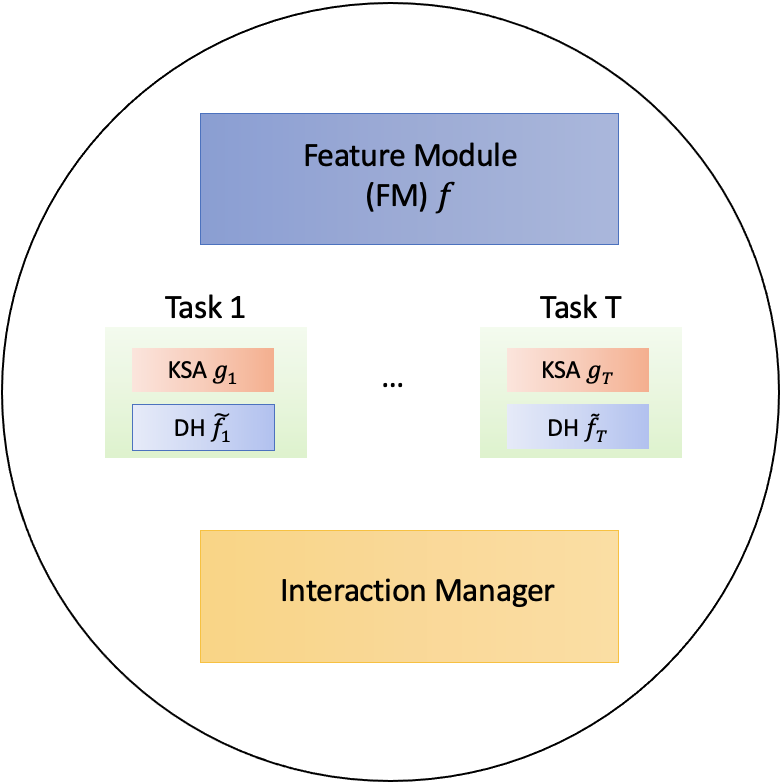}
    \caption{Each LENC node contains the following modules: a) a Feature Module (FM), which incorporates a DNN for feature extraction, b) a set of $T$ Decision Heads (DHs), each one with the corresponding Knowledge Self-Assessment (KSA) module attached, and c) the Interaction Manager (IM), which regulates the communications with other LENC nodes. A LENC community is composed of multiple LENC nodes.}
    \label{fig:node}
\end{figure}

Fig. \ref{fig:node} represents the structure of a LENC node. Each LENC node contains a Feature Module (FM), namely a DNN model $f$, parameterized by $\mathbf{w}_s$, which is typically a DNN. $f$ is assumed to be shared across $T$ tasks, $T \geq 0$, on which it has been trained using the appropriate corresponding training datasets $\mathcal{D}_\tau, \tau = 1,\dots, T$. The shared FM culminates in $T$ individual Decision Heads (DHs) $\Tilde{f}_\tau, \tau=1,\dots, T$, parameterized by $\mathbf{w}_\tau$, so that the decision for each task is taken by the function $y_\tau = \Tilde{f}_\tau(f(\mathbf{x};\mathbf{w}_s);\mathbf{w}_\tau), \tau=1,\dots, T$ for an input test data point $\mathbf{x}$. This structure allows the deployed node to support multiple tasks, which potentially have a different number of semantic classes, using a single DNN. Optionally, $a_\tau$ can be stored along with $\Tilde{f}_\tau$: it is this node's known accuracy in the test set of $\mathcal{D}_\tau$, as measured before deployment using any task-appropriate evaluation metric.

Each node also contains $T$ Knowledge Self-Assessment (KSA) modules $g_1, ..., g_T$, which assess if incoming test data points match the distribution of the corresponding original training datasets $\mathcal{D}_\tau$. Each LENC node can autonomously decide to act as either a student or a teacher DNN, depending on the output of its KSAs. Furthermore, each node contains a set of Interaction Rules (IRs) that specify its communications with other nodes and, thus, shape teacher-student interactions.

If a LENC node's KSA modules indicate that it does not have sufficient knowledge of current test inputs, this particular node can temporarily become a student and trigger a search for one or more teacher nodes in the LENC community. Once they are found, the student may learn from them using its IRs. This process, when triggered by a specific incoming test data stream that proves to be unknown or not well-known, is called an \textit{education cycle}: one iteration of KSA-triggered teacher discovery followed by policy-based knowledge transfer.

The various LENC node modules are subsequently detailed.
\vspace{-1em}
\subsubsection{Knowledge Self-Assessment (KSA) Module}
\label{sssec:KSAModule}
The $\tau$-th KSA module of a LENC node consists of an OOD detector $g_\tau(\mathbf{x})$ corresponding to the $\tau$-th task, $\tau=1,\dots, T$, the node has been trained on. It is assumed that each OOD detector has been pretrained/preconfigured per node and per task on that task’s ID training data, prior to deployment. During system deployment, it assigns an OOD score to a stream of incoming test data points $\mathbf{x} \in \mathcal{X}_\tau $, thus assessing the relevant knowledge of the FM. Thus, the $T$ node's KSA modules are tailored to the $T$ known training datasets $D_\tau$ that the node has encountered before its deployment. In short, the KSA modules provide each node with task-agnostic, on-line knowledge self-assessment capabilities, allowing it to assess on its own the DH most knowledgeable/relevant to the current test data point $\mathbf{x}$. As a result, during FM inference on $\mathbf{x}$, the node automatically detects and activates/utilizes the $j$-th DH $\Tilde{f}_j, j = 1, \dots, T$ as the one most relevant to $\mathbf{x}$, since $j = \arg \min (g_1, \dots, g_T)$ is the index of the supported task with known training data most similar to the ones found in $\mathcal{X}$. This implies that, assuming the decision for the $\tau$-th task is taken among $c_\tau$ classes, then the node will finally predict $y_j = \Tilde{f}_j(f(\mathbf{x};\mathbf{w}_s);\mathbf{w}_j)$.

Given an unlabelled data point $\mathbf{x}$ incoming from the external world, each of the node's KSA modules may output one of three potential verdicts, based on the respective raw $g_\tau$ value: i) the task is not known at all by the FM (non-expert), ii) the task is known, but the node's relevant knowledge is limited (limited knowledge), or iii) the task is well-known by the node's FM (expert). In the first and the second case, an education cycle will be triggered as a response (see Section \ref{ssec::learning}), but in the limited existing knowledge scenario the existing DH will be employed for receiving education, instead of appending a new DH. In the third case, no education is kick-started and the node only infers on the incoming data stream. The verdict depends on the KSA's internally computed OOD score for the current stream and on two relevant manually prespecified (and, in principle, task-specific) thresholds: $\delta$ and $\epsilon$. Thus, the first, second, or third case is activated if $g_\tau(\mathbf{x}) \geq \delta$, $ \epsilon < g_\tau(\mathbf{x}) < \delta$, or $g_\tau(\mathbf{x}) \leq \epsilon$, respectively.

\subsubsection{Interaction Rules}
\label{ssec:Method_IRs}
The LENC node Interaction Rules (IRs) are defined by the LENC framework and serve three basic LENC node interaction functions. The first one is that they specify the interaction between a deployed LENC node and the external environment, which can constantly fetch unlabelled test input data points for analysis in the form of a data stream $\mathcal{D}^s = \left\{\mathbf{x}_i\right\}$, where $i=0, \dots, M^s$ and $M^s > 0$ is the total number of currently received data points. If the node's KSA modules respond that the FM does not know the current test data distribution well or at all, and is therefore a non-expert or has limited knowledge, an education cycle is triggered: DNN experts are automatically searched for within the LENC node community, in order to serve as teachers.

The second IR function specifies the transmission of the data stream $\mathcal{D}^s$ from LENC node $s$ to the other $N-1$ LENC community participants and the reception of their responses: $\left\{q_n, n = 1,\ldots, N, n \neq s\right\}$. The response $q_n$ from the $n$-th LENC node is set to 0 if none of that node's KSA modules recognizes $\mathcal{D}^s$ as in-distribution. Otherwise, it is a non-zero numerical score that denotes how well the $n$-th node and its most suitable DH know the distribution of data $\mathcal{D}^s$. Assuming that the KSA modules of the $n$-th node indicate that the latter's $j$-th DH/supported task is the most suitable to $\mathcal{D}^s$, three different policies can be \textit{alternatively} employed for computing $q_n$. In turn, $q_n$ can be used for \emph{teacher selection} at each knowledge transaction. These IRs are shown in Algorithm \ref{alg: interaction_rules}. The three alternative policies are:
\begin{itemize}
\item[a)] $q_n = a^n_j$, where $a^n_j$ is the optionally prestored average classification \textit{accuracy} (e.g., Correct Classification Rate).
\item[b)] $q_n$ can be a function $g_j$ of an \textit{OOD score} internally computed by the $j$-th KSA module of the $n$-th node, given $\mathcal{D}^s$: $q_n = g_j$, where higher $g_j$ implies worse match\footnote{In the particular LENC implementation evaluated in this article, the OOD score is computed according to Eq. (\ref{eq: lr_calculation}), i.e., the LR method.}.
\item[c)] $q_n$ can be a scalar measure of the \textit{disagreement} between the current student LENC node and the $n$-th LENC node. To this end, the \emph{churn} metric \cite{Fard2016} can be used: the misclassification rate of student predictions given as input $\mathcal{D}^s$, using the corresponding predictions of the $n$-th node as pseudo-ground-truth. For a set of $v$ coupled predictions $\mathbf{y}^n = \left[y^n_1,...,y^n_v\right]$ and $\mathbf{y}^s = \left[y^s_1,...,y^s_v\right]$, this is defined as:
\begin{equation}
q_n = C\left(\mathbf{y}^n, \mathbf{y}^s\right) = \frac{1}{v}\sum_{i=1}^v r(y^n_i, y^s_i),    
\end{equation} 
\begin{equation}
r(y^n_i, y^s_i) = \begin{cases}
    1, & y^n_i \neq y^s_i \\
    0, & y^n_i = y^s_i.
    \end{cases} 
\end{equation} 
\end{itemize}

The LENC node with the best non-zero response ($t = \arg \max(\{q_u, u\neq s\})$ for options a) and c), $t = \arg \min(\{q_u, u\neq s\})$ for option b)) is then selected for transferring its knowledge of $\mathcal{D}^s$ to the querying student LENC node. \hl{In the case of Disagreement Policy (c), this will lead to learning from the teacher LENC node having the maximal output disagreement with the student LENC node output, among KSA-approved candidate teachers. This policy is intended to prioritize task-relevant peers that carry information not yet matched by the student; however, disagreement is a heuristic and not a correctness guarantee. A confidently wrong teacher could still induce negative transfer, so safety-critical deployments should combine this policy with confidence/OOD gates, teacher-history scores, or multi-teacher agreement.}

Third, the IRs are responsible for specifying the actual teacher-student knowledge exchanges. Various policies can be alternatively employed, resulting in messages of different content. The four different knowledge transfer policies implemented in LENC are extensively discussed in Section \ref{ssec::learning}.

\begin{algorithm}
\caption{LENC Node Interaction Rules for Teacher Selection.}
\label{alg: interaction_rules}
\begin{algorithmic}[1]
\State \textbf{Input:} Unlabelled test data stream $\mathcal{D}_s = \{x_i\}_{i=1}^{M_s}$
\For{each incoming data point $x$}
    \State KSA: compute OOD score $g_{\tau}(x)$
    \If{$g_{\tau}(x) \geq \delta$} \Comment{Unknown task}
        \State Trigger education cycle: Search for teacher nodes
    \ElsIf{$\epsilon < g_{\tau}(x) < \delta$} \Comment{Limited knowledge}
        \State Use existing Decision Head (DH) for education
    \Else \Comment{Task is well-known}
        \State Proceed with inference only
    \EndIf
\EndFor
\State \textbf{Teacher Selection:}
\For{each candidate node $n \neq s$}
    \State Compute response score $q_n$:
    \State \hspace{0.5cm} a) $q_n = a_{nj}$ (Average classification accuracy)
    \State \hspace{0.5cm} b) $q_n = g_j(\mathcal{D}_s)$ (OOD score function)
    \State \hspace{0.5cm} c) $q_n =$ Disagreement-based score using churn metric
\EndFor
\State Select teacher node $t = \arg\max q_n$ \Comment{Or $\arg\min q_n$ for option b)}
\end{algorithmic}
\end{algorithm}

\vspace{-1em}
\subsection{Learning a Novel Task}
\label{ssec::learning}
\begin{figure*}
    \centering
    \begin{subfigure}[b]{0.45\textwidth}
        \includegraphics[width=\textwidth]{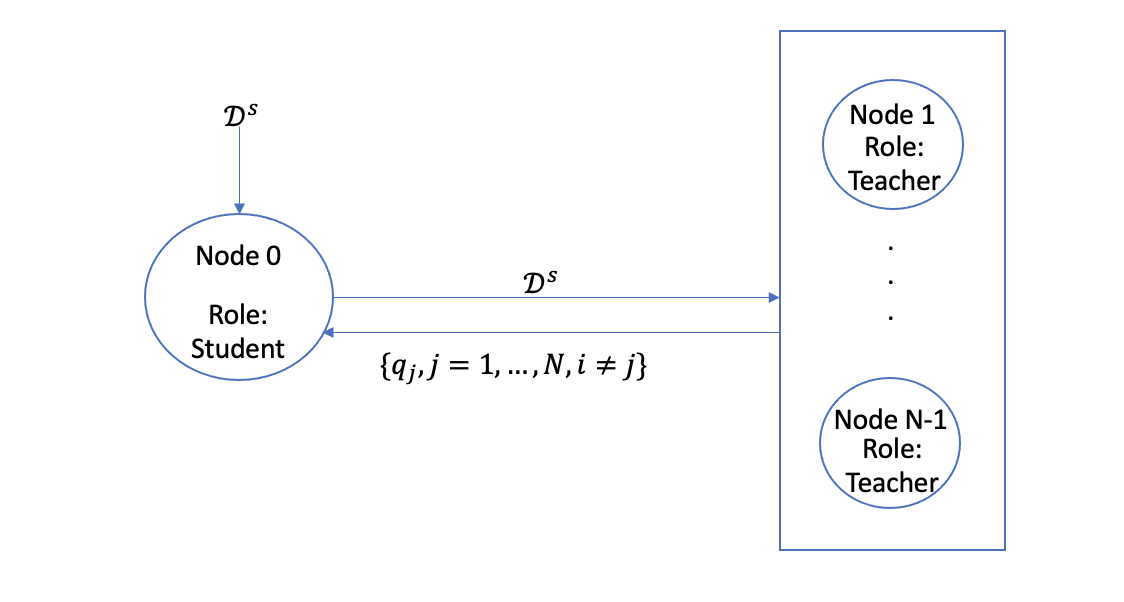}
        \caption{}
        \label{fig:int_sub1}
    \end{subfigure}
    \hfill
    \begin{subfigure}[b]{0.45\textwidth}
        \includegraphics[width=\textwidth]{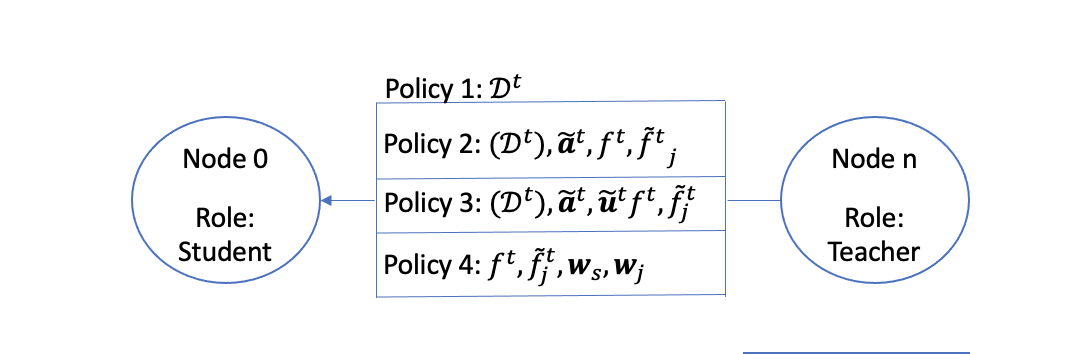}
        \caption{}
        \label{fig:int_sub2}
    \end{subfigure}
    \caption{The LENC inter-node interactions: a) second IR function, b) third IR function. In a), the incoming data stream is transmitted to all LENC nodes, so that \hl{a suitable teacher can be selected} according to teacher selection rules. In b), knowledge is transferred from teacher to student using the appropriate policy (determined via the policy selection rules).}
    \label{fig:interactions}
\end{figure*}

As in the case of human societies, the external world constantly provides novel data stimuli \hl{to be classified} by one or more of the deployed individual LENC nodes participating in the LENC community. For each such test input data stream $\mathcal{D}^s$, the first question that needs to be answered is if the triggered $s$-th LENC node is knowledgeable of it. Such a question is answered using this LENC node's KSA modules. If the task is judged to be known and the LENC node is considered an expert (see Section \ref{sssec:KSAModule}), no action is taken by LENC and the node proceeds to infer its own predictions for $\mathcal{D}^s$. In any other case, an education cycle is triggered: the $s$-th LENC node temporarily assumes a student role and sends $\mathcal{D}^s$ to other active LENC nodes within the community. Each of the other $N-1$ LENC nodes receives $\mathcal{D}^s$ and forwards it through its own KSA modules. Then, the $n$-th node replies to the student with $q_n$, $n \in \{1,\dots,N\}$\textbackslash$\{s\}$, which specifies whether it knows the distribution of $\mathcal{D}^s$ or not. From this point on, in the context of this particular knowledge transaction, any LENC node with non-zero $q_n$ is considered a potential teacher. In the implemented LENC system the student node automatically selects as actual teacher the LENC node with the highest $q_n$ score (see Fig. \ref{fig:interactions}). However, in principle, knowledge transfer from multiple teachers to a simple student LENC node can also be supported.

Assuming that the student LENC node already knows $T^s \geq 0$ tasks, using $T^s$ existing DHs $\Tilde{f}^s_1, \dots, \Tilde{f}^s_{T^s}$, it will now learn the new task driven by the dataset $\mathcal{D}^s$, using the selected teacher's knowledge. If the student LENC node has limited prior knowledge of the task (see Section \ref{sssec:KSAModule}), the existing DH for the current task will be enhanced via knowledge transfer from the teacher. In contrast, if the student has no prior knowledge of the task, a new DH will be generated and appended. As hinted in Section \ref{ssec:Method_IRs}, the LENC framework offers four alternative policies for teacher-student knowledge transfer. Whatever policy is selected, it can be combined with a CL method, so that the student retains its previously acquired knowledge and does not experience catastrophic forgetting, just as in human society. The four \emph{knowledge transfer policies} are illustrated in Fig. \ref{fig:interactions}, demonstrated in Algorithm \ref{alg: transfer_policies} and explained below.\\

    

        

\textbf{Policy 1: Training Data Transfer Policy}
The student LENC node receives from the selected teacher LENC node the latter's stored labeled training dataset $\mathcal{D}^t_j, j\in{1,\dots,T}$, which was utilized before deployment to train the teacher for its $j$-th task. This known dataset is the most relevant to the current test data stream $\mathcal{D}^s$. The student LENC node trains on it in a classical manner, using the selected CL method. Of course, if the student LENC node was previously untrained ($T^s = 0$) then the integrated CL algorithm does not need to be activated. Any common problem-specific loss function can be employed. A simple example for classification follows below:
\begin{equation}
\mathcal{L}_t = \begin{cases}
    \mathcal{L}_c  + \alpha \mathcal{L}_h(\mathbf{y}, \Tilde{\mathbf{y}}^s), & T^s \geq 1 \\
    \alpha \mathcal{L}_h(\mathbf{y}, \Tilde{\mathbf{y}}^s) , & T^s = 0, \\
\end{cases}
\label{eq::KnowledgeTransfer1}
\end{equation}
\noindent where $\Tilde{\mathbf{y}}^s$ is the student node's prediction for input in $\mathcal{D}^t_j$, $\mathbf{y}$ is the respective one-hot-encoded ground-truth label from $\mathcal{D}^t_j$, $\mathcal{L}_h(\cdot,\cdot)$ is the cross-entropy loss, $\mathcal{L}_c$ is the EWC CL regularizer and $\alpha$ is a hyperparameter controlling the relative influence of the main cross-entropy loss term.
\begin{algorithm}
\caption{LENC Knowledge Transfer Policies}
\label{alg: transfer_policies}
\begin{algorithmic}[1]
\State \textbf{Policy 1: Training Data Transfer}

\State Receive labeled dataset $\mathcal{D}_j^t$ from teacher
\State Train using standard CL loss:
\[
L_t =
\begin{cases}
L_c + \alpha L_h(y, \tilde{y}^s), & T^s \geq 1 \\
\alpha L_h(y, \tilde{y}^s), & T^s = 0
\end{cases}
\]

\State \textbf{Policy 2: Knowledge Distillation (KD)}
\State Receive soft activations $\tilde{a}_j^t$ from teacher:
\[
\tilde{a}_j^t = f_j^t(f(x_t; w_s); w_j)
\]
\State Optimize KL divergence loss:
\[
L_d=
\begin{cases}
    L_c + \alpha L_h(y, \tilde{y}^s) + \beta KL(\tilde{a}_j^t, \tilde{a}^s), & T^s \geq 1 \\
    \alpha L_h(y, \tilde{y}^s) + \beta KL(\tilde{a}_j^t, \tilde{a}^s), & T^s = 0
    \end{cases}
\]

\State \textbf{Policy 3: Intermediate Feature Distillation}
\If{Teacher and Student have common architecture}
    \State Receive intermediate layer activations $\tilde{u}_t$ and soft activations $\tilde{a}_j^t$
    \State Train using FitNet loss:
    \[
    L_f = \begin{cases}
    L_c + \alpha L_h(y, \tilde{y}^s) + \beta KL(\tilde{a}_j^t, \tilde{a}^s) \\ 
    + \gamma \sum ||\tilde{u}_t - \tilde{u}_s||, & T^s \geq 1 \\
    \alpha L_h(y, \tilde{y}^s) + \beta KL(\tilde{a}_j^t, \tilde{a}^s) \\ 
    + \gamma \sum ||\tilde{u}_t - \tilde{u}_s||, & T^s = 0
    \end{cases}
\]
    
\EndIf

\State 
\textbf{Policy 4: DNN Model Transfer}
\If{Student is untrained $(T^s = 0)$}
    \State Receive full model parameters $(w_s^t, w_j^t)$ from teacher
    \State Copy teacher's Feature Module (FM) and Decision Head (DH)
\EndIf
\end{algorithmic}
\end{algorithm}

\textbf{Policies 2-3: Knowledge Distillation Policies}
When the second or third policy is selected, the student receives from the selected teacher node the latter's soft-output activations $\mathbf{\Tilde{a}}^t_j = \Tilde{f}^t_j(f^t(\mathbf{x}^t;\mathbf{w}_s);\mathbf{w}_j)$. Note that, obviously, for classification problems the teacher's respective prediction is $\Tilde{y}_j^t = \arg \max (\mathbf{\Tilde{a}}^t_j)$. There are two alternative options for selecting which inputs these teacher activations are generated from: $\mathbf{x}^t$ can come either from $\mathcal{D}^s$ (unknown ground-truth labels) or from $\mathcal{D}^t_j$ (known ground-truth labels). With the first option, the student is subsequently trained using $\mathcal{D}^s$ and a KD loss (e.g., the one from \cite{hinton2015distilling}, for simple classification problems). With the second input option, the teacher sends to the student $\mathcal{D}^t_j$ as well and the latter is trained by a combination of a KD loss and a common, problem-specific loss (as in Policy 1). CL is utilized with both options, if the student is not previously untrained, as in Policy 1. A simple classification loss function example of the second input option without CL is the following one:
\begin{equation}
\mathcal{L}_d =  \mathcal{L}_h + \beta KL(\mathbf{\Tilde{a}}^t_j, \mathbf{\Tilde{a}}^s),
\end{equation}
\noindent where \textit{KL} denotes the Kullback-Leibler (KL) divergence and $\beta$ is a hyperparameter to control the relative influence of the KD loss.

Policy 3 is similar to Policy 2, with the exception that the teacher also sends its intermediate layer activations $\mathbf{\Tilde{u}}^t = \left\{\mathbf{\Tilde{u}}^t_1, \dots, \mathbf{\Tilde{u}}^t_k \right\}$, where $k$ denotes a neural layer index within the teacher's FM. The intermediate layers of interest must be predefined for each DNN architecture, while this policy is currently only available when the FMs of the teacher and the student share a common FM architecture. In this case, knowledge transfer to the student proceeds by combining a KD loss term with the Fit-Net \cite{romero2014fitnets} loss, designed to integrate the knowledge of intermediate teacher node layers. Thus, a simple example loss function for Policy 3 in the case of classification problems, using the first input option (without CL) is the following one:
\begin{equation}
\mathcal{L}_f = KL(\mathbf{\Tilde{a}}^t, \mathbf{\Tilde{a}}^s) + \gamma \sum \| \Tilde{\mathbf{u}}^t - \Tilde{\mathbf{u}}^s \|, 
\end{equation}
\noindent where $\gamma$ is a KD hyperparameter to control the relative influence of the Fit-Net loss.\\

\textbf{Policy 4: DNN Model Transfer}\\
In Policy 4 the teacher node sends to the student LENC node its FM and DH parameters ($\mathbf{w}^t_s$, $\mathbf{w}^t_j$) and structure ($f^t$ and $\Tilde{f}^t_j$), for its $j$-th task, $j\in \{1, \dots, T^t\}$. This option is only applicable when the student LENC node is fully untrained at the beginning of the current knowledge transaction ($T^s = 0$). Essentially, the student LENC node is transformed into a copy of the teacher LENC node.\\

\vspace{-1em}
\textbf{Policy Selection Rules.} 
The four different knowledge transfer policies support the LENC functionalities under different conditions. Thus, a set of rules has been devised for automatic selection of the current policy based on the external environment within which the LENC community operates, as defined by the user. These user-set conditions are essentially answers to the following three questions:
\begin{itemize}
    \item[a)] Are there model architecture/dataset/parameter privacy limitations during knowledge exchanges?
    \item[b)] Are there network traffic limitations within the LENC community?
    \item[c)] Is there a need to minimize the latency of each knowledge transaction?
\end{itemize}

For example, privacy and network traffic considerations may lead to limitations on whether each node's known training datasets, internal parameters, or architectural details can be shared. Under the strongest sharing limitations, where restrictions are placed simultaneously on architectures, datasets and model parameters, only Policy 2 with the first input option (distill the teacher's soft-output activations, given $\mathcal{D}^s$ as input) can be actually employed. This is the most privacy-preserving and architecture/dataset-agnostic option, thus it is the default choice.

If it is known that the student and the teacher share a common neural architecture, and there are no bandwidth restrictions, then Policy 3 is selected as a better option due to a higher degree of student guidance by the teacher. For both Policy 2 and Policy 3, the second input option (distill the teacher's soft-output and/or intermediate activations, given $\mathcal{D}^t_j$ as input) is selected only if there are no dataset privacy and/or network traffic restrictions since $\mathcal{D}^t_j$ is likely significantly larger than the current test batch $\mathcal{D}^s$. Policy 4 is selected as a training-free option only if the following conditions are concurrently true:
\begin{itemize}
    \item[a)] Each knowledge transfer needs to happen instantly due to latency restrictions,
    \item[b)] There are no model architecture/parameter privacy limitations, and
    \item[c)] The student is previously untrained.
\end{itemize}
Finally, Policy 1 is activated only when the following conditions hold simultaneously: 
\begin{itemize}
    \item[a)] There are neither dataset privacy nor network traffic limitations,
    \item[b)] There are no model architecture privacy restrictions (thus the student can know/learn the teacher's architecture),
    \item[c)] The student's architecture is significantly more complex than the teacher one.
\end{itemize}

Obviously, Policy 1 and the second input option for Policies 2-3 assume that each LENC node's original training dataset for each of its supported tasks is stored along with its FM and DHs. If this is not the case, only Policy 4 and the first input option for Policies 2-3 are applicable.
\vspace{-0.9em}
\section{Experimental Evaluation}
Existing CKD literature has several limitations, as it leaves important use-cases unaddressed. Most CKD methods do not consider the potential help of experts who have acquired their knowledge asynchronously with their peers, or cases where the external environment provides only a few unlabelled data points to the node community. Such scenarios resemble more closely human learning in real communities. To evaluate the effectiveness of the LENC framework in similar setups, it was experimentally compared against existing CKD methods in scenarios involving on-line unlabelled CKD from a pretrained teacher.

In order to facilitate direct and fair comparisons with competing CKD algorithms, the main experiments were conducted using Policy 2 with the first input option for knowledge transfer (see Section \ref{ssec::learning}) and the Disagreement Policy for teacher selection (see Section \ref{ssec:Method_IRs} for details on $q_n$). The Disagreement Policy does not require cross-node comparability of raw OOD magnitudes.

Two different sets of main experiments were performed: a) whole-image classification with untrained nodes and one expert, where no CL capabilities need to be activated, and b) whole-image classification where most nodes contain prior knowledge. In the first case, a traditional CKD setup is simulated (up to a degree), in order to facilitate comparisons against competing CKD methods. Thus there is only a single task ($T^s=0$), with only a single node having been pretrained and able to serve as an expert teacher. However, in contrast to existing CKD methods, this fact is not a priori known to the students, but automatically discovered by the LENC framework. In the second case, LENC is configured to run in a setup that demonstrates a fuller extent of its true capabilities. Most of the participating nodes are pretrained in different ways and LENC showcases its ability to handle on-line learning of multiple tasks on-the-fly, based on incoming unlabelled data points.

CL is employed only in the second experimental setup, while the KSA modules are utilized in both. However, in the CKD experiments the KSA modules are only used for identifying which node is a potential teacher, while in the CL experiments they additionally address task-agnostic CL by automatically identifying the task index (as described in Section \ref{sec::Method}). The baseline CL method of EWC \cite{kirkpatrick2017overcoming} was selected for integration into the implemented LENC system, due to its combination of simplicity and good performance. These qualities of EWC underlie its continuing use as a building block by newer CL approaches \cite{zhang2023slca, heng2024selective}. Similarly, LR \cite{xiao2020likelihood} was selected for OOD detection within the KSA modules.

Additionally, comprehensive ablation studies were performed. These experiments were designed to showcase the LENC framework's  good performance in on-line unlabelled CKD, the ability of LENC nodes to avoid catastrophic forgetting without access to the task boundaries, the effectiveness of the different knowledge transfer policies and the different teacher selection policies (for computing $q_n$), LENC's CKD performance across varying architectures and data stream sizes, as well as LENC's performance when key components experience performance degradation.

\vspace{-1em}
\subsection{CKD Experimental Setup}
Following common CKD evaluation protocols \cite{zhang2018deep, guo2020online, hou2017dualnet, zhu2018knowledge, wu2021peer, zhang2022weighted, qian2022switchable}, the LENC framework is evaluated on datasets CIFAR-10 and CIFAR-100, using nodes with the neural architectures ResNet \cite{he2016deep}, Wide-ResNet (WRN) \cite{zagoruyko2016wide} and VGG \cite{simonyan2014very}. A pretrained ResNet-18 is employed as the only teacher, while two alternative sizes are utilized for the incoming data stream $\mathcal{D}^s$ that originates in the external environment: 1000 and 5000 data points. The use of a pretrained expert excludes from the comparisons CKD methods for collaborative learning from scratch with neural branches of identical architecture \cite{zhu2018knowledge, wu2021peer, zhang2022weighted}. The data points of a stream $\mathcal{D}^s$ are randomly sampled from the teacher's actual training dataset, with 10 different $\mathcal{D}^s$ sets constructed in this manner. Each student receives sequentially the 10 streams, with each one triggering an education cycle; although the node is no longer entirely untrained after the first cycle, it is not an expert either. The competing CKD methods were adapted to distill the teacher's response, instead of training with ground-truth, in order to enable fair comparisons with LENC. Although the competing SwitOKD method \cite{qian2022switchable} uses ground-truth labels to calculate the teacher's influence, it is also included in the evaluation.

The LENC community included two homogeneous students (2 ResNet-18 models) and two heterogeneous students (WRN-16-4 and VGG11). After hyperparameter search, the batch size was set to 128 and Stochastic Gradient Descent (SGD) was adopted as an optimizer, with an initial learning rate of 1e-3 and momentum of 0.9. The number of epochs for knowledge transfer was set to 100. Table \ref{tab:comparison} reports average community accuracy in the respective test set for the last education cycle, over all students and across 10 independent runs with different random seeds, along with the standard deviation over the different runs. \hl{A likely reason for LENC’s higher performance in this setup is the employed teacher selection policy: given the lack of ground-truth annotation, each student node picks a KSA-approved candidate teacher with which it disagrees the most, to leverage diverse knowledge within the community.} Instead, the adapted competing CKD methods also consider the non-expert responses of other nodes.

\begin{table*}[ht]
\centering
\caption{Comparisons of LENC with competing CKD methods, for incoming data streams $\mathcal{D}^s$ of sizes 1000 and 5000. The reported metrics are the mean and std deviation of test set accuracy (\%) over 10 independent runs, where each run averages accuracy over student nodes after the last education cycle.}
\label{tab:comparison}
\begin{tabular}{lcccccc}
\toprule
\textbf{Dataset} & \textbf{Students} & \textbf{Stream Size} & \textbf{DML} & \textbf{KDCL} & \textbf{SwitOKD}  & \textbf{LENC (proposed)} \\
\midrule
\multirow{4}{*}{CIFAR-10} & ResNet-18 \& ResNet-18 & \multirow{2}{*}{1000} & 52.20$\pm$0.52 & 62.23$\pm$0.15 & 56.15$\pm$0.73 & \textbf{76.93$\pm$0.71} \\
 & WRN-16-4 \& VGG11 &  & 51.17$\pm$0.71 & 62.09 $\pm$ 0.21 & 57.85$\pm$0.80 & \textbf{70.16$\pm$0.82} \\
 
 \addlinespace[0.5ex]
\cline{3-7}
\addlinespace[1ex]
 & ResNet-18 \& ResNet-18 & \multirow{2}{*}{5000} & 77.85$\pm$0.31 & 85.76$\pm$0.07 & 79.08$\pm$0.70 & \textbf{86.31$\pm$ 0.32} \\
 & WRN-16-4 \& VGG11 &  & 75.56$\pm$0.82 & 84.47 $\pm$ 0.08 & 78.79$\pm$0.68 & \textbf{87.12$\pm$0.24} \\
 
\midrule
\multirow{4}{*}{CIFAR-100} &ResNet-18 \& ResNet-18 & \multirow{2}{*}{1000} & 9.77$\pm$0.25 & 25.16$\pm$0.12 & 13.71$\pm$0.57 & \textbf{34.96$\pm$0.47} \\
 & WRN-16-4 \& VGG11 &  & 6.12$\pm$0.38 & 27.59$\pm$0.19 & 14.72$\pm$0.61 & \textbf{29.75$\pm$0.49} \\
 
 \addlinespace[0.5ex]
\cline{3-7}
\addlinespace[1ex]
 & ResNet-18 \& ResNet-18 & \multirow{2}{*}{5000} & 31.53$\pm$0.31 & 58.70$\pm$0.09 &  35.31$\pm$0.29 & \textbf{65.02$\pm$0.13} \\
 & WRN-16-4 \& VGG11 &  & 8.30$\pm$0.16 & 56.94$\pm$0.12 & 37.27$\pm$0.45 & \textbf{58.18$\pm$0.17} \\

\bottomrule
\end{tabular}

\end{table*}

\begin{table}[h]
\centering
\caption{Average test accuracy (\%) and std deviation on Tiny-ImageNet, using a ViT-base teacher and two ViT-small students. LENC is compared against its best CKD competitor.}
\label{tab:tinyimagenet}
\begin{tabular}{lcc}
\toprule
\textbf{Stream Size} & \textbf{KDCL} & \textbf{LENC (proposed)} \\
\midrule
1000 & \textbf{10.11$\pm$0.17} & 10.05$\pm$0.08 \\
5000 & 9.98$\pm$0.10 & \textbf{10.08$\pm$0.13} \\
10000 & \textbf{10.24$\pm$0.08} & 10.16$\pm$0.11 \\
20000 & 10.31$\pm$0.19 & \textbf{17.25$\pm$0.36} \\
\bottomrule
\end{tabular}
\end{table}

A variant of this unlabelled experiment was also conducted using the more challenging Tiny-ImageNet dataset \cite{le2015tiny} and Transformer DNNs; this resembles a hard real-world scenario. The LENC community was formed by a ViT-base model as the teacher, achieving on its own an accuracy of 86.35\% on the task, and two ViT-small students. In this setup, LENC was compared only against the best competitor \cite{guo2020online}, according to the outcomes of previous experiments. The data points of $\mathcal{D}^s$ were randomly sampled from the teacher's actual training dataset, with 15 different $\mathcal{D}^s$ sets constructed in this manner. Results indicate that LENC does not clearly outperform the competing method for small $\mathcal{D}^s$ sizes. However, as shown in Table \ref{tab:tinyimagenet}, by gradually increasing the size of $\mathcal{D}^s$ and repeating the experiment, at a size of 20000 data stream points LENC surpasses the best competitor by a large margin.
\vspace{-1em}
\subsection{Continual Learning Experiments}
The LENC framework can operate in a completely task-agnostic nature: the current $\mathcal{D}^s$ data stream is sent to potential teachers to self-assess their knowledge and reply accordingly if they are aware of the current task. This enables the community to include multiple experts from multiple tasks, by providing an autonomous mechanism for finding suitable teachers, while concurrently supporting pretrained students via CL. The ability of the nodes to self-assess their knowledge, via their KSA modules, enables them to identify the task at hand and augment their knowledge of it. This is not feasible at all in existing CKD methods and, therefore, could not be evaluated comparatively to competing CKD algorithms. Instead, a dedicated CL experimental evaluation was conducted. 

The experimental evaluation protocol for task-agnostic CL follows the relevant literature \cite{zhu2022tame, kirichenko2021task} and evaluates the LENC framework on the SPLIT-MNIST, SPLIT-CIFAR-10, and SPLIT-CIFAR-100 datasets. Each dataset is split into 5 subsets/tasks containing 2, 2, and 20 classes respectively, resulting in 15 independent tasks overall. This experiment is a variant of the previous one. The community contains 15 pretrained nodes, each one knowing one of the 15 tasks, and 3 initially untrained nodes that will necessarily act as students. Each incoming data stream $\mathcal{D}^s$, given to a student by the external environment, is of size 1000 and concerns 1 of the 15 tasks. A different $\mathcal{D}^s$ is randomly sampled and transmitted to a student for each of the 15 tasks, sequentially. Each specific student receives 5 streams $\mathcal{D}^s$ corresponding to the 5 different tasks of a single dataset (i.e., one dataset ``assigned" to each of the 3 students). The different $\mathcal{D}^s$ streams arrive in a random order and trigger consecutive education cycles: the students autonomously identify their need for education and find \hl{a suitable} teacher within the LENC community, regardless of the total number of nodes or the tasks they are aware of. As a result, each of the 3 different student nodes is consecutively educated on the 5 CL tasks corresponding to one of the 3 datasets, without any information on the task indices. For these experiments, the batch size was set to 128 and SGD was adopted as an optimizer with an initial learning rate of 1e-3 and momentum of 0.9. ResNet-18 was used as a FM and the \emph{EWC} hyperparameter $\lambda$ was set to 500.

\begin{figure*}
    \begin{subfigure}{0.5\textwidth}
        \centering
        \begin{tikzpicture}
            \begin{axis}[
                width=0.75\textwidth,
                height=5cm,
                xlabel={Education Cycles per Task},
                ylabel={Task Accuracy (\%)},
                xmin=0, xmax=25,
                ymin=95, ymax=100,
                legend style={at={(1.05,1)}, anchor=north west},
                legend cell align={left},
                cycle list name=color list,
                mark options={solid},
                grid=major,
            ]
            
            \addplot[mark=*, color=blue] table[x=LC, y=Task 0, col sep=comma] {MNIST-SPLIT_per.csv};
            \addlegendentry{Task 0}
            
            \addplot[mark=*, color=red] table[x=LC, y=Task 1, col sep=comma] {MNIST-SPLIT_per.csv};
            \addlegendentry{Task 1}
            
            \addplot[mark=square*, color=green] table[x=LC, y=Task 2, col sep=comma] {MNIST-SPLIT_per.csv};
            \addlegendentry{Task 2}
            
            \addplot[mark=square*, color=magenta] table[x=LC, y=Task 3, col sep=comma] {MNIST-SPLIT_per.csv};
            \addlegendentry{Task 3}
            
            \addplot[mark=triangle*, color=cyan] table[x=LC, y=Task 4, col sep=comma] {MNIST-SPLIT_per.csv};
            \addlegendentry{Task 4}
            
            \end{axis}
        \end{tikzpicture}
        \caption{MNIST-SPLIT.}
        \label{fig:mnist_taskvalues}
    \end{subfigure}
    \begin{subfigure}{0.5\textwidth}
        \centering
        \begin{tikzpicture}
            \begin{axis}[
                width=0.75\textwidth,
                height=5cm,
                xlabel={Education Cycles per Task},
                xmin=0, xmax=25,
                ymin=50, ymax=85,
                legend style={at={(1.05,1)}, anchor=north west},
                legend cell align={left},
                cycle list name=color list,
                mark options={solid},
                grid=major,
            ]
            
            \addplot[mark=*, color=blue] table[x=LC, y=Task 0, col sep=comma] {CIFAR-SPLIT10_per.csv};
            \addlegendentry{Task 0}
            
            \addplot[mark=*, color=red] table[x=LC, y=Task 1, col sep=comma] {CIFAR-SPLIT10_per.csv};
            \addlegendentry{Task 1}
            
            \addplot[mark=square*, color=green] table[x=LC, y=Task 2, col sep=comma] {CIFAR-SPLIT10_per.csv};
            \addlegendentry{Task 2}
            
            \addplot[mark=square*, color=magenta] table[x=LC, y=Task 3, col sep=comma] {CIFAR-SPLIT10_per.csv};
            \addlegendentry{Task 3}
            
            \addplot[mark=triangle*, color=cyan] table[x=LC, y=Task 4, col sep=comma] {CIFAR-SPLIT10_per.csv};
            \addlegendentry{Task 4}
            
            \end{axis}
        \end{tikzpicture}
        \caption{CIFAR-10-SPLIT.}
        \label{fig:cifar_taskvalues}
    \end{subfigure}
    
    \begin{subfigure}{\textwidth}
        \centering
        \begin{tikzpicture}
            \begin{axis}[
                width=0.4\textwidth,
                height=5cm,
                xlabel={Education Cycles per Task},
                ylabel={Task Accuracy (\%)},
                xmin=0, xmax=25,
                ymin=5, ymax=35,
                legend style={at={(1.05,1)}, anchor=north west},
                legend cell align={left},
                cycle list name=color list,
                mark options={solid},
                grid=major,
            ]
            
            \addplot[mark=*, color=blue] table[x=LC, y=Task 0, col sep=comma] {CIFAR-SPLIT100_per.csv};
            \addlegendentry{Task 0}
            
            \addplot[mark=*, color=red] table[x=LC, y=Task 1, col sep=comma] {CIFAR-SPLIT100_per.csv};
            \addlegendentry{Task 1}
            
            \addplot[mark=square*, color=green] table[x=LC, y=Task 2, col sep=comma] {CIFAR-SPLIT100_per.csv};
            \addlegendentry{Task 2}
            
            \addplot[mark=square*, color=magenta] table[x=LC, y=Task 3, col sep=comma] {CIFAR-SPLIT100_per.csv};
            \addlegendentry{Task 3}
            
            \addplot[mark=triangle*, color=cyan] table[x=LC, y=Task 4, col sep=comma] {CIFAR-SPLIT100_per.csv};
            \addlegendentry{Task 4}
            
            \end{axis}
        \end{tikzpicture}
        \caption{CIFAR-100 SPLIT.}
        \label{fig:cifar_taskvalues_multiplied}
    \end{subfigure}
    \caption{CL experiments for SPLIT-MNIST, SPLIT-CIFAR-10 and SPLIT-CIFAR-100. Best seen in color.}
    \label{fig:continual learning}
\end{figure*}

The results are visible in Fig. \ref{fig:continual learning}. The final average accuracy (mean of the five per-task test accuracies evaluated after learning all five tasks) is $99.18\%, 68.62\%$ and $28.35\%$ for SPLIT-MNIST, SPLIT-CIFAR-10, and SPLIT-CIFAR-100 respectively. Although these results decline compared to the results in Table \ref{tab:comparison} for the LENC framework (ResNet-18 with 1000 data points), it is crucial to notice that the method can achieve CL and adaptation with only a few randomly sampled batches.
\vspace{-1em}
\subsection{Ablation Studies}
Ablation studies were performed using the main experimental setup, to show whether picking the correct teacher improves performance in an on-line unlabelled CKD setting. It is demonstrated how the size of $\mathcal{D}^s$ affects all compared methods. Additionally, the performance of the LENC framework for varying architectures is examined. Subsequently, the relative influence of the number of education cycles and of participating nodes is studied. Finally, alternative LENC knowledge transfer and teacher selection policies are assessed, given that the main experiments only rely on Policy 2 with the first input option (for knowledge transfer) and the Disagreement Policy (for teacher selection). Details follow below.

\begin{figure*}
    \centering
    \begin{tikzpicture}
        \centering
        \begin{axis}[
            width=0.6\textwidth,
            height=5cm,
            xlabel={Number of data points},
            ylabel={Average Accuracy (\%)},
            ymin=0, ymax=100,
            legend style={at={(1.05,1)}, anchor=north west},
            legend cell align={left},
            cycle list name=color list,
            mark options={solid},
            grid=major,
        ]
        
        \addplot[mark=diamond*, color=purple] table[x=NS, y=LENC, col sep=comma] {abl_res_master.csv};
        \addlegendentry{LENC (ours)}
        
        \addplot[mark=x, color=black] table[x=NS, y=KDCL, col sep=comma] {abl_res_master.csv};
        \addlegendentry{KDCL}
        
        \addplot[mark=square*, color=magenta] table[x=NS, y=SwitOKD, col sep=comma] {abl_res_master.csv};
        \addlegendentry{SwitOKD}

        \addplot[mark=triangle*, color=green] table[x=NS, y=DML, col sep=comma] {abl_res_master.csv};
        \addlegendentry{DML}

        \end{axis}
    \end{tikzpicture}
    \caption{Average student accuracy (\%) for varying $\mathcal{D}^s$ sizes in the CIFAR-10 dataset. Best seen in color.}
    \label{fig:abl_samples}
\end{figure*}

\subsubsection{Varying data stream sizes}
Fig. \ref{fig:abl_samples} shows the performance of the compared methods for varying $\mathcal{D}^s$ sizes. All the CKD methods were evaluated for incoming streams of 50, 100, 200, 500, 1000, 2000, and 5000 data points, to comparatively validate the on-line learning capabilities of the LENC framework. LENC proves to be the most tolerant to small batch sizes, thus showcasing its usability in an important real-world use-case: when a node faces unknown current input data and needs to acquire relevant knowledge as soon as possible, in order to respond immediately. DML presents a minor advantage of 1\% for data stream size of 50 samples, however it does not scale this well for the rest of the experiments, showcasing an overall lower performance compared to LENC. The latter's superiority most likely stems from its avoidance of fusing the participating nodes' responses. Instead, it uses a simple, yet effective way of evaluating all available nodes for tutoring and picking the correct teacher at each education cycle.

\begin{table*}[ht]
\centering
\caption{Multiple neural architectures ablation study. For each of the four groups, the final test accuracy (\%) is reported for all student nodes. The teachers are not trained during the process (-).}
\label{tab:ablation architectures}
\begin{tabular}{llcccc}
\toprule
\textbf{Teacher} & \textbf{Stream Size} & \textbf{ResNet-18} & \textbf{VGG11} & \textbf{WRN-16-4} & \textbf{ViT}  \\
\midrule
\multirow{6}{*}{ResNet-18} & 500 & - & 63.35 & 42.26 & -  \\
  & 1000 & - & 64.07 & 32.31 & -   \\
  & 5000 & - & 75.62 & 63.18 & -   \\
 \addlinespace[0.5ex]
\cline{2-6}
\addlinespace[1ex]
  & 500 & - & 60.23 & 42.08 & 44.87  \\
  & 1000 & -  & 68.57 & 50.15 & 50.10 \\
  & 5000 & -  & 83.62 & 79.62 & 64.47 \\
\midrule
   \multirow{3}{*}{ViT} & 500 & 39.40 & - & 57.12 & - \\
    & 1000 & 35.90 & - & 58.61 & - \\
    & 5000 & 60.20 & - & 74.36 & - \\
 
 \addlinespace[0.5ex]
\cline{1-6}
\addlinespace[1ex]
 \multirow{3}{*}{WRN-16-4} & 500 & 62.10 & 63.18 & - & 48.75 \\
 & 1000 & 71.29 & 70.49 & - & 54.47 \\
 & 5000 & 69.30 & 78.80 & - & 69.3 \\

\bottomrule
\end{tabular}

\end{table*}
\vspace{-0.1em}
\subsubsection{Diverse neural architectures}
The use of the Disagreement Policy for teacher selection, using the churn metric, motivated an exploration of the role of diversity in CKD environments. Thus, multiple architectures are independently evaluated for $\mathcal{D}^s$ streams of size 500, 1000, and 5000, per education cycle. Following relevant literature \cite{zhu2018knowledge, wu2021peer, zhang2022weighted}, the neural architectures ResNet-18, WRN-16-4, and VGG11 were employed, along with a more recent Vision Transformer (ViT) \cite{dosovitskiy2020image}. In this experimental setup, there are 3 expert nodes and 4 different groups of nodes. The experts are a ResNet-18, a ViT, and a WRN-16-4, pretrained on the CIFAR-10 dataset with a test accuracy of 92.31\%, 80.91\%, and 77.94 \%, respectively. Varying node groups are used, composed of 2-3 students per expert, to investigate how the architectures affect LENC-powered CKD. The results are demonstrated in Table \ref{tab:ablation architectures}. As it can be seen, the simple presence of ViT within the community, without changing at all the teacher and the other peer nodes, boosts the performance of VGG11 and WRN-16-4. Given that ViT does not perform well on small datasets \cite{zhu2023understanding}, this result further validates the argument that diversity in training enhances the overall knowledge of the community: diversity allows the other peers to achieve higher test accuracy. Another remark on Table \ref{tab:ablation architectures} is that ViT proves to be a more effective teacher for WRN-16-4 than ResNet-18.

\begin{figure*}[ht]
    \centering
    \begin{subfigure}[b]{0.4\textwidth}
        \centering
        \begin{tikzpicture}
            \begin{axis}[
                width=\textwidth,
                height=5cm,
                xlabel={Number of education cycles.},
                ylabel={Average Accuracy (\%)},
                ymin=0, ymax=100,
                legend style={at={(1.05,1)}, anchor=north west},
                legend cell align={left},
                cycle list name=color list,
                mark options={solid},
                grid=major,
            ]
            \addplot[mark=*, color=black] table[x=TC, y=Value, col sep=comma] {TC_abl.csv};
            \addlegendentry{Node Average}
            
            \end{axis}
        \end{tikzpicture}
        \caption{Average test accuracy for different numbers of education cycles.}
        \label{fig:ablation_teachingcircles}
    \end{subfigure}
    \hfill
    \begin{subfigure}[b]{0.4\textwidth}
        \centering
        \begin{tikzpicture}
            \begin{axis}[
                width=\textwidth,
                height=5cm,
                xlabel={Number of Nodes},
                ymin=0, ymax=100,
                cycle list name=color list,
                mark options={solid},
                grid=major,
            ]
            \addplot[mark=*, color=black] table[x=TC, y=Value, col sep=comma] {A_abl.csv};
            
            \end{axis}
        \end{tikzpicture}
        \caption{Average test accuracy for different numbers of total nodes in the community.}
        \label{fig:abl_nodes}
    \end{subfigure}
    \caption{Ablation studies for different total numbers of nodes and education cycles.}
    \label{fig:ablation_tc_nn}
\end{figure*}

\subsubsection{Scalability studies}
Fig. \ref{fig:ablation_tc_nn} demonstrates the relative impact of the total number of education cycles and of participating nodes on the LENC framework. As expected, more education cycles lead to higher community performance (see Fig. \ref{fig:ablation_teachingcircles}), assuming a fixed number of nodes (1 teacher and 3 students were used here). Similarly, Fig. \ref{fig:abl_nodes} depicts the average test accuracy as the total number of nodes in the community rises, while keeping the number of education cycles fixed to 5. The performance reaches a peak for 3 nodes (2 students and 1 teacher). Empirical investigation of the results indicates that as the number of nodes increases, the number of education cycles should also increase accordingly in order to retain stable performance. In these experiments, $\mathcal{D}^s$ streams of size 1000 were randomly sampled and transmitted from the external environment at the start of each education cycle.

\subsubsection{Alternative knowledge transfer policies}
Besides Policy 2 with the first input option for knowledge transfer, other LENC knowledge transfer policies were also evaluated and compared in separate, complementary experiments, for on-line unlabelled CKD using a pretrained teacher. \hl{The CIFAR-10 dataset was used, while a pretrained ResNet-18 with a test classification accuracy of 91.97\% was deployed as a teacher model.} Two initially untrained ResNet-18 models were deployed as student LENC nodes. Policy 1 is equivalent to classic training, if the LENC node is initially untrained and teacher selection has been completed. For Policies 2-3, both input options were independently evaluated: a) the first one using the unlabelled dataset $\mathcal{D}^s$, and b) the second one using the labeled dataset $\mathcal{D}^t_j$.

All experiments were conducted for a total of 10 education cycles. Table \ref{tab:policies_comparison} reports the average student LENC node accuracy after the final education cycle, for both input options and for data streams of varying size, randomly sampled at each education cycle, in the unlabelled case. As it can be seen, the first input option for Policies 2-3 (using $\mathcal{D}^s$) can lead to very good performance with only minimal network traffic overhead, compared to the second one. Moreover, Policy 2 outperforms Policy 1 when using the second input option and even approaches it in classification accuracy when using the (lightweight) first input option.

\begin{table}[ht]
\centering
\caption{Comparisons of the LENC knowledge transfer policies, for incoming data streams $\mathcal{D}^s$ of sizes 100, 500, 1000, 5000, and 60000 (full dataset). Policies 2-3 are independently evaluated with both unlabelled (using $\mathcal{D}^s$) and labeled (using $\mathcal{D}^t_j$) input options. The average test classification accuracy (\%) of the student LENC nodes is reported.}
\label{tab:policies_comparison}
\begin{tabular}{lcccc}
\toprule
\textbf{Dataset} & \textbf{Stream Size} & \textbf{Policy 1} & \textbf{Policy 2} & \textbf{Policy 3}  \\
\midrule
$\mathcal{D}^t_j$ & 60000 & 91.97 & \textbf{93.72} & 93.59 \\

\midrule
\multirow{5}{*}{$\mathcal{D}^s$} 
  & 60000 & - & 91.86 & \textbf{92.07} \\
 
\cline{2-5}
  & 100 & - & \textbf{37.75} & 37.11 \\
  
  & 500 & - & 61.13 &  \textbf{62.48} \\
  
  & 1000 & - & 74.04 & \textbf{74.29} \\
  
  & 5000 & - & \textbf{90.15} &  90.05 \\

\bottomrule
\end{tabular}

\end{table}

\begin{figure*}
    \centering
    \begin{tikzpicture}
        \centering
        \begin{axis}[
            width=0.4\textwidth,
            height=3cm,
            xlabel={Number of data points},
            ylabel={Average Accuracy (\%)},
            ymin=0, ymax=100,
            legend style={at={(1.05,1)}, anchor=north west},
            legend cell align={left},
            cycle list name=color list,
            mark options={solid},
            grid=major,
        ]
        
        \addplot[mark=star, color=orange] table[x=NS, y=LENC(acc), col sep=comma] {master_selection_options.csv};
        \addlegendentry{LENC (acc)}

        \addplot[mark=diamond*, color=purple] table[x=NS, y=LENC(div), col sep=comma] {master_selection_options.csv};
        \addlegendentry{LENC (div)}

        \addplot[mark=*, color=cyan] table[x=NS, y=LENC(ood), col sep=comma] {master_selection_options.csv};
        \addlegendentry{LENC (ood)}
        
        \addplot[mark=x, color=black] table[x=NS, y=KDCL, col sep=comma] {master_selection_options.csv};
        \addlegendentry{KDCL}
        
        \addplot[mark=square*, color=magenta] table[x=NS, y=SwitOKD, col sep=comma] {master_selection_options.csv};
        \addlegendentry{SwitOKD}

        \addplot[mark=triangle*, color=green] table[x=NS, y=DML, col sep=comma] {master_selection_options.csv};
        \addlegendentry{DML}
        \end{axis}
    \end{tikzpicture}
    \caption{Average student LENC node classification accuracy (\%) for varying $\mathcal{D}^s$ sizes in the CIFAR-10 dataset. The 3 alternative LENC teacher selection policies are compared against competing methods. Best seen in color.}
    \label{fig:qn_score_samples}
\end{figure*}
\begin{figure}
    \centering
    \begin{tikzpicture}
    \begin{axis}[
        xlabel={$\lambda$ hyperparameter},
        ylabel={Accuracy (\%)},
        xmin=100, xmax=2000,
        ymin=50, ymax=60,
        xtick={0, 500, 1000, 2000},
        ytick={50, 52, 54, 56, 58, 60},
        legend pos=north east,
        grid=major,
        width=6cm, height=4cm,
        cycle list name=color list
    ]
    
        \addplot[mark=star, color=orange] table[x=NS, y=0, col sep=comma] {CL_master.csv};
        \addlegendentry{Task 0}

        \addplot[mark=diamond*, color=purple] table[x=NS, y=1, col sep=comma] {CL_master.csv};
        \addlegendentry{Task 1}

        \addplot[mark=*, color=cyan] table[x=NS, y=2, col sep=comma] {CL_master.csv};
        \addlegendentry{Task 2}
        
        \addplot[mark=x, color=black] table[x=NS, y=3, col sep=comma] {CL_master.csv};
        \addlegendentry{Task 3}
        
        \addplot[mark=square*, color=magenta] table[x=NS, y=4, col sep=comma] {CL_master.csv};
        \addlegendentry{Task 4}

        \end{axis}
    \end{tikzpicture}
    \caption{Classification accuracy on CIFAR100-SPLIT tasks after learning all 5 tasks, using different values for the CL hyperparameter $\lambda$. Best seen in color.}
    \label{fig:lambda_ablation}

\end{figure}

\begin{figure}[ht]
    \centering
   
        \begin{tikzpicture}
            \begin{axis}[
                width=0.4\textwidth,
                height=5cm,
                xlabel={Binary Noise on the KSA module output},
                ylabel={Average Accuracy (\%)},
                ymin=0, ymax=100,
                legend cell align={left},
                cycle list name=color list,
                mark options={solid},
                grid=major,
            ]
            \addplot[mark=*, color=red] table[x=noise, y=avg, col sep=comma] {ood-noise-master.csv};
            \addlegendentry{Node Average}
            
            \end{axis}
        \end{tikzpicture}

    \caption{KSA module robustness analysis by adding binary noise to the KSA modules' output.}
    \label{fig:ablation_ood_noise}
\end{figure}
\vspace{-0.3em}
\subsubsection{Alternative teacher selection policies}
Regarding the teacher selection policies (see Section \ref{ssec:Method_IRs} for details on computing $q_n$), the LENC framework was evaluated on CIFAR-10 using LENC nodes with ResNet \cite{he2016deep}, Wide-ResNet (WRN) \cite{zagoruyko2016wide} and VGG \cite{simonyan2014very} neural architectures. A pretrained ResNet-18 model is deployed as a potential teacher, while the untrained students continuously receive data streams $\mathcal{D}^s$ with varying cardinality, ranging from 50 to 5000 data points. Regarding knowledge transfer, the default Policy 2 with the first input option was used here. A student receives sequentially the 10 streams, with each one triggering an education cycle; although this LENC node is no longer entirely untrained after the first cycle, it is not an expert either. The teacher node is available every two education cycles for tutoring, so that the students can capture the diverse knowledge of their peers. Fig. \ref{fig:qn_score_samples} illustrates the average student LENC node classification accuracy after 10 education cycles. As it can be seen, the LENC framework outperforms the state-of-the-art methods for any choice of teacher selection policy among the three possible ones. Between them, the Accuracy Policy leads to the highest performance. However, this policy relies on static, pre-stored accuracy scores, which are not available for the initially untrained LENC nodes; this prevents them from ever acting as teachers, after enough education cycles.
\vspace{-0.3em}
\subsubsection{LENC robustness}
A set of ablation studies were dedicated to assess the robustness of the LENC framework against degradation of its critical internal components. First, robustness to suboptimal performance of the integrated CL method was evaluated by artificially tuning the regularizer hyperparameter $\lambda$ for EWC (see Section \ref{ssec:CL}) to various non-optimal values. Fig. \ref{fig:lambda_ablation} displays the results on the CIFAR100-SPLIT dataset, by training using only raw data for different values of $\lambda \in \left\{100, 200, 500, 1000, 2000\right\}$. All experiments used streams of 5000 data points per task for training. As expected, higher values of $\lambda$ generally lead to increased final accuracy for the two tasks that are learned first (Task 0 and Task 1). However, the highest average classification accuracy over all tasks was 55.72\% for $\lambda=200$.

Similarly, LENC's robustness against suboptimal performance of its internal OOD detector was also measured in the on-line setting, where few incoming data points are available, using the integrated LR method (see Section \ref{ssec: OOD}). The conducted experiment involved independently training the OOD detector with varying number of data points from CIFAR-10, ranging from 50 to 5000, and then querying it on unknown images of CIFAR-10 and SVHN. The experiment indicated that the implemented version of LENC does not work correctly when its OOD detectors are trained with less than 1000 data points, due to failure of its KSA modules to accurately distinguish between known and unknown data.

LENC's robustness to OOD failures was also evaluated by repeating the CKD experiment of \cref{tab:comparison} in the scenario of two untrained ResNet-18 students. The data streams contained 1000 CIFAR-10 data points. In this case, the KSA outputs of all nodes were corrupted by various levels of binary noise (to transform ``non-expert" verdicts to ``expert" verdicts, and vice versa), at a probability range of 0.1 to 0.5. This emulates the behavior of suboptimal OOD detectors, implying partial KSA failure. LENC demonstrated high tolerance to this, as shown in Figure \ref{fig:ablation_ood_noise}, with only a slight decrease in the average accuracy of LENC nodes upon increasing binary noise.

LENC's robustness on noisy data was also assessed, by repeating the CKD experiment of \cref{tab:comparison} in the scenario of two untrained ResNet-18 students. The data streams contained 1000 and 5000 CIFAR-10 and CIFAR-100 data points, now corrupted by 2D Gaussian noise of standard deviation equal to 0.01/0.05/0.2, for low/medium/hard noise level, respectively. In the CIFAR-10/5000 case, the students learned the task at hand via LENC with an average test accuracy score of 86.17\%/83.93\%/16.04\%, respectively. As seen in \cref{tab:comparison}, the corresponding accuracy for clean data (zero noise) is 86.31\%. Thus, the students failed the challenge of hard corruption, but proved robust at low and medium noise levels. Additional results are provided in the Supplementary Material.

\vspace{-1em}
\subsection{LENC failure modes}

The ablation study revealed certain limitations of LENC. Given that KSA modules rely on OOD detectors and are responsible for a variety of LENC tasks, it is unsurprising that many of said limitations stem from unexpected KSA module behavior, likely due to OOD failure. In such a scenario, the most important issue arising is wrong teacher selection, i.e., a teacher is misidentified as an expert for the current task. In this case, the current education cycle is not positively contributing to the student's knowledge. However, due to multiple education cycles, this situation is not irreversible and the student can end up learning the correct task by identifying the appropriate teacher in the next education cycles. Second, if a node already knows a task and the relevant KSA module marks the current data stream as OOD, then the node will append a redundant/duplicate DH, with a new KSA attached to it, for the same task. However, this is not fatal for the node's operation: during the future education cycles, the duplicate DH/KSA might evolve to recognize such data streams as ID and overcome this issue automatically. The experimental results indicate that the KSA OOD detectors trained with too few data points are not accurate enough. However, KSA training with more than 1000 data points leads to sufficient accuracy. Finally, the LENC framework might be susceptible to highly corrupted data, as shown by the relevant robustness experiments.

\hl{Of course, LENC assumes complementary expertise in the community, where at least one peer covers the incoming stream. If all peers’ KSAs indicate non-expertise, LENC can detect this case but cannot distill missing knowledge without labels or an expert. In practice, such no-teacher events should trigger abstention, buffering, active annotation, or later onboarding of a new expert.}
\vspace{-1.2em}
\subsection{Discussion}
\hl{The current LENC framework has limitations.} First, scalability to larger communities is primarily constrained by teacher discovery: each education cycle broadcasts streams to all peers and triggers per-node KSA plus inference to compute $q_n$. Thus,  compute and network size grow roughly linearly and the few best teachers attract a disproportionately large share of requests, creating an imbalance that overloads those nodes, raising latency and contention. Communication is the next limiter, with knowledge transfer Policy 3 being orders-of-magnitude heavier than Policy 2. Each student consults only one teacher per cycle, so adding more nodes does not increase how much knowledge is fused/disseminated at each iteration. But since each cycle still broadcasts/collects across multiple heterogeneous peers, frequent cycles inflate per-round memory, bandwidth, and latency. Potential remedies include hierarchical clustering of nodes, selective knowledge routing, or communication compression strategies.

Another open challenge concerns robustness to adversarial or noisy teacher responses: Byzantine nodes or corrupted feedback could mislead students and destabilize the community. Future extensions could incorporate Byzantine-resilient aggregation, trust-weighted teacher selection, or consensus-based filtering to ensure resilience in decentralized and adversarial environments.

\hl{A related limitation is the fully open-world case where no peer has relevant expertise. In such a case, LENC should abstain from distillation, flag or buffer the unsupported stream, and optionally cluster repeated unsupported streams for later active annotation, self-supervised representation adaptation, or onboarding of a newly trained expert node. Once such knowledge is created, it can be registered as a new DH/KSA pair and disseminated through ordinary LENC education cycles. Thus, handling no-expert streams requires an outer open-world discovery/annotation layer, rather than a different teacher-selection rule alone.}

\hl{Details relevant to a potential real-world implementation are in the Supplementary Material.} In addition, the manual thresholds $\delta$ and $\epsilon$ employed by the KSA modules may be insufficient under concept drift or evolving data distributions. Dynamic strategies for threshold adjustment, e.g., on-line calibration, reinforcement learning, or meta-learning approaches, could provide more adaptive task-confidence decisions for distinguishing between non-expert, limited-knowledge, and expert roles. Addressing these issues will enhance the robustness and scalability of LENC, while moving it closer to viable, real-world deployment.
\vspace{-0.2em}
\section{Conclusions}
\label{sec::Conclusions}
The Learning-by-Education Node Community (LENC) framework emulates human‐community learning for multi‐node, on-line Collaborative Knowledge Distillation (CKD) with DNNs and unlabelled data. Nodes autonomously adopt teacher or student roles, using no information on the task boundaries or index, to handle diverse data distributions, using Continual Learning (CL) and Out-of-Distribution (OOD) detection to learn on-the-fly from \hl{a peer deemed as suitable}. Proof-of-concept image classification experiments show that LENC leverages peer diversity to achieve state-of-the-art on-line unlabelled CKD performance while supporting task-agnostic CL. Among multi-node CKD protocols, LENC uniquely combines CKD with task-agnostic, on-line CL from unlabelled data and dynamic role assignment, bringing CKD closer to human communities. \hl{Practical deployments may include edge, robotic, automotive, or IoT communities where deployed nodes have complementary expertise and need to exchange distilled knowledge under bandwidth, privacy, or latency constraints. Fully open-world operation, where no peer has relevant expertise, remains outside the present CKD formulation and requires external novelty-discovery, annotation, or expert-onboarding mechanisms.}

\vspace{-0.8em}
\bibliography{refs}
\bibliographystyle{IEEEtran}
\begin{IEEEbiography}
[{\includegraphics[width=1in,height=1.25in,clip,keepaspectratio]{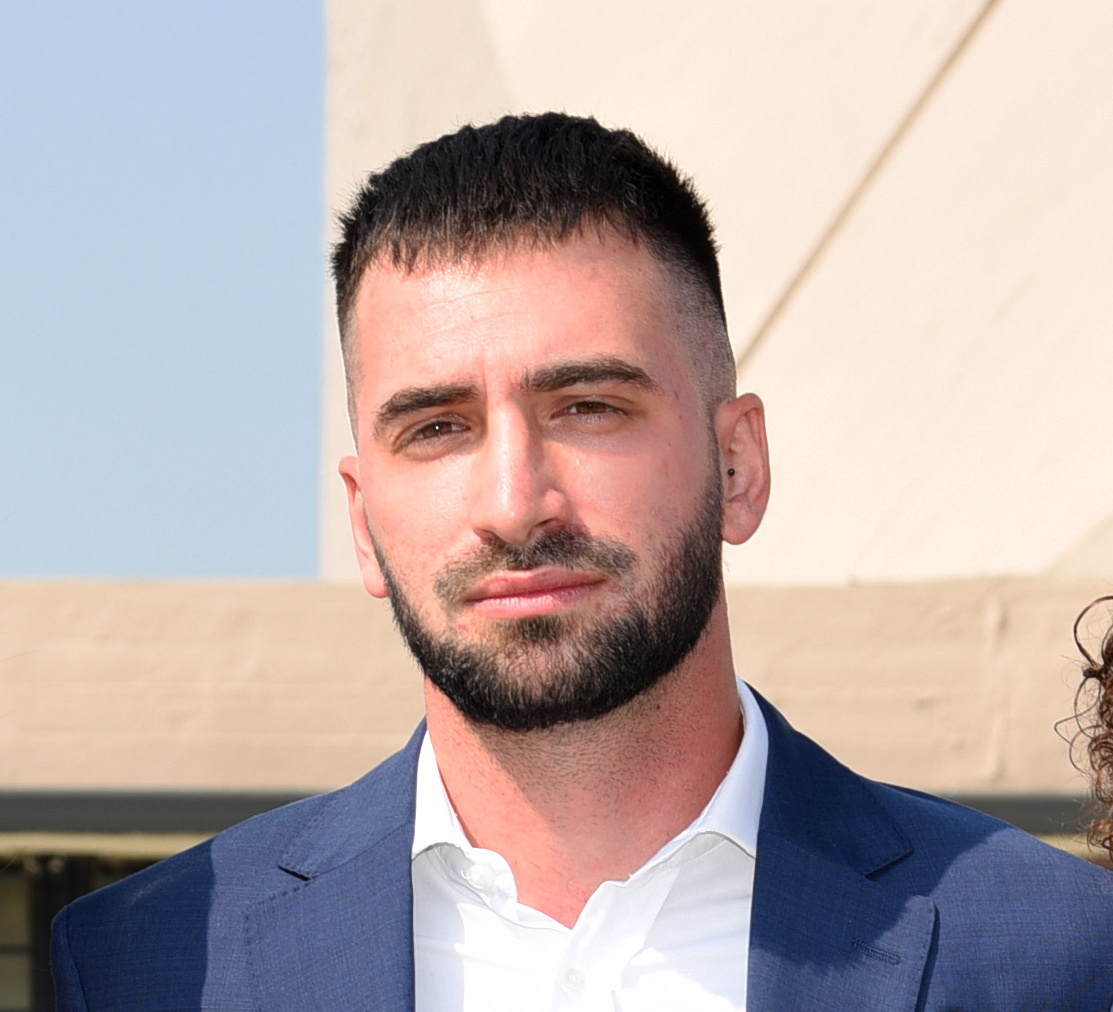}}] {Anestis Kaimakamidis} received a M.Eng. in electrical and computer engineering (2023) from the Aristotle University of Thessaloniki (AUTH), Greece. He was a research assistant with the Artificial Intelligence and Information Analysis Laboratory of AUTH, specializing in artificial intelligence. He is currently pursuing a M.Sc. in artificial intelligence at the Northeastern University of Boston, U.S.A. His research interests include computer vision, machine learning and continual learning.
\end{IEEEbiography}

\begin{IEEEbiography}[{\includegraphics[width=1in,height=1.25in,clip,keepaspectratio]{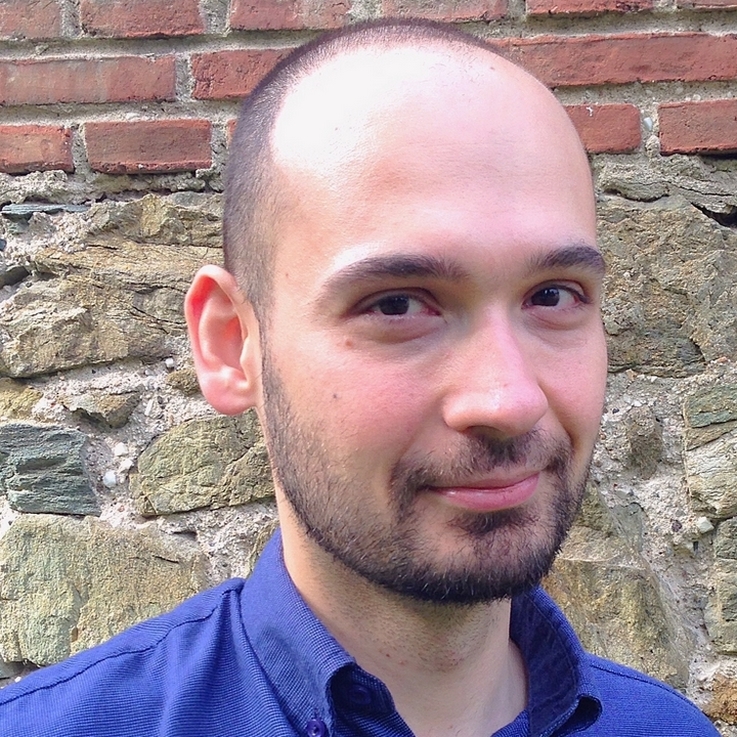}}] {Dr. Ioannis Mademlis} (S’17-M’18-SM’22) is a computer scientist specializing in computer vision and machine learning. He received his Ph.D. in 2018 from the Aristotle University of Thessaloniki (AUTH), in Thessaloniki, Greece. He was a postdoctoral research associate at AUTH and at the Harokopio University of Athens, in Athens, Greece. He has been an adjunct lecturer at the Athens University of Economics and Business and at the National and Kapodistrian University of Athens, in Athens, Greece. He has participated in 6 European Union-funded R\&I projects, having (co-)authored approximately 80 publications in academic journals and international conferences. He is a member of the AI Curriculum Committee of the International Artificial Intelligence Doctoral Academy (AIDA). His current research interests include machine learning, computer vision, autonomous robotics and human-computer interaction.
\end{IEEEbiography}

\begin{IEEEbiography}[{\includegraphics[width=1in,height=1.25in,clip,keepaspectratio]{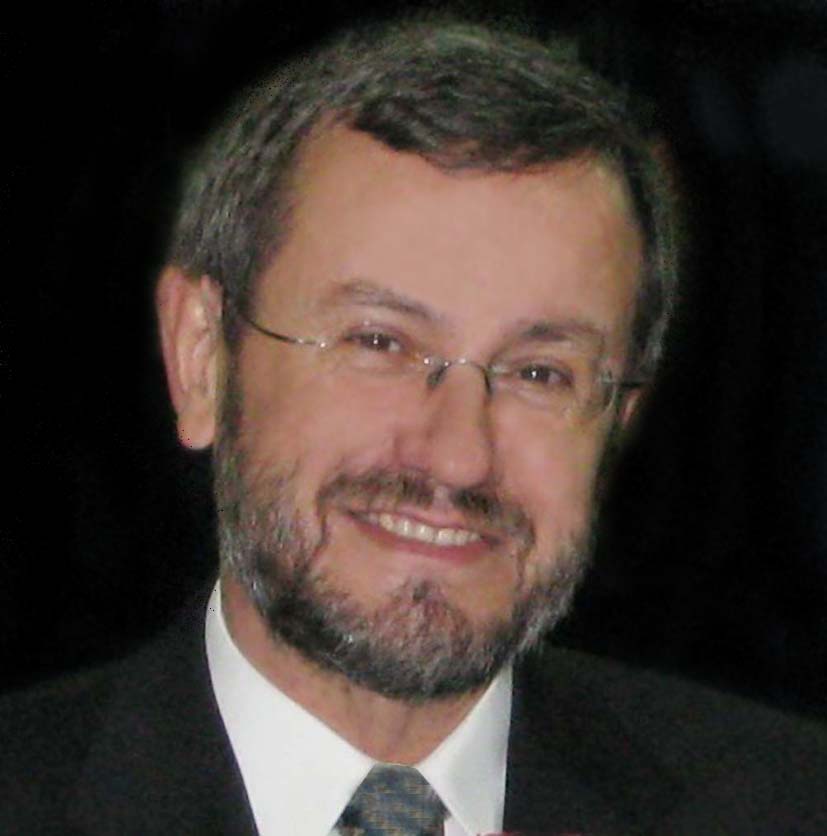}}] {Prof. {Ioannis Pitas}} (SM’94-F’07, IEEE Fellow, IEEE Distinguished Lecturer, EURASIP Fellow) received the Diploma and Ph.D. degree in electrical engineering, in 1980 and 1985, respectively, from the Aristotle University of Thessaloniki (AUTH), Greece. Since 1994, he has been a Professor at the Department of Informatics of AUTH and Director of the Artificial Intelligence and Information Analysis (AIIA) lab. He served as a Visiting Professor at several Universities. His current interests are in the areas of computer vision, machine learning, autonomous systems, intelligent digital media, image/video processing, human-centred interfaces, affective computing, 3D imaging and biomedical imaging. He has published over 900 papers, contributed in 47 books in his areas of interest and edited or (co-)authored another 11 books. He has also been member of the program committee of many scientific conferences and workshops. In the past he served as Associate Editor or co-Editor of 9 international journals and General or Technical Chair of 4 international conferences. He participated in 70 R\&D projects, primarily funded by the European Union and is/was principal investigator/researcher in 42 such projects. He has 31600+ citations to his work and h-index 92+ (Google Scholar). He leads the International AI Doctoral Academy (AIDA) and coordinates the HE project “TEMA” (g.a.n. 101093003).
\end{IEEEbiography}

\end{document}